\documentclass{article}

 \usepackage[preprint]{neurips_2025}

\usepackage[utf8]{inputenc} % allow utf-8 input
\usepackage[T1]{fontenc}    % use 8-bit T1 fonts
\usepackage{hyperref}       % hyperlinks
\usepackage{url}            % simple URL typesetting
\usepackage{booktabs}       % professional-quality tables
\usepackage{amsfonts}       % blackboard math symbols
\usepackage{nicefrac}       % compact symbols for 1/2, etc.
\usepackage{microtype}      % microtypography
\usepackage{xcolor}         % colors
\usepackage{graphicx}
\usepackage{subcaption}
\usepackage{dsfont}
\usepackage{amsmath}
\usepackage{amssymb}
\usepackage{cleveref}
\usepackage{pifont}
\newcommand{\cmark}{\text{\ding{51}}}
\newcommand{\xmark}{\text{\ding{55}}}

\title{TRIBE: TRImodal Brain Encoder \\for whole-brain fMRI response prediction}

\author{%
  Stéphane d'Ascoli \\
  Meta AI \\
  \texttt{sdascoli@meta.com} \\
  \And
  Jérémy Rapin \\
  Meta AI \\
  \texttt{jrapin@meta.com}
  \And
  Yohann Benchetrit \\
  Meta AI\\
  \texttt{ybenchetrit@meta.com} \\
  \And
  Hubert Banville\\
  Meta AI \\
  \texttt{hubertjb@meta.com} \\
  \And 
  Jean-Rémi King\\
  Meta AI \\
  \texttt{jeanremi@meta.com} \\
}

\begin{document}

\maketitle

\begin{abstract}
% Motivation
Historically, neuroscience has progressed by fragmenting into specialized domains, each focusing on isolated modalities, tasks, or brain regions. While fruitful, this approach hinders the development of a unified model of cognition. Here, we introduce TRIBE, the first deep neural network trained to predict brain responses to stimuli across multiple modalities, cortical areas and individuals. By combining the pretrained representations of text, audio and video foundational models and handling their time-evolving nature with a transformer, our model can precisely model the spatial and temporal fMRI responses to videos, achieving the first place in the Algonauts 2025 brain encoding competition with a significant margin over competitors. Ablations show that while unimodal models can reliably predict their corresponding cortical networks (e.g. visual or auditory networks), they are systematically outperformed by our multimodal model in high-level associative cortices. Currently applied to perception and comprehension, our approach paves the way towards building an integrative model of representations in the human brain. Our code is available at \url{https://github.com/facebookresearch/algonauts-2025}.
% Building accurate models of how the brain responds to stimuli has become a central question in neuroscience. 
% SOTA
% Existing approaches typically use linear models which take as input the intermediate representations of a deep neural network. 
% Challenge
% However, these approaches suffer from two main shortcomings: (i) the linear mappings are limited in expressive power, and (ii) they fail to capture the multimodal nature of the stimuli.
% Approach
% In this work, we introduce the first deep-learning based method for predicting brain responses to multimodal stimuli. 
% Our model, TRIBE, combines features from state-of-the art text, audio and video models and handles their time-evolving nature by leveraging transformers.
% Result
% TRIBE achieves the first place in the ongoing Algonauts 2025 competition and achieves encoding scores near to the noise ceiling in some areas.
% We show that the three modalities typically dominate in the expected brain regions associated to audition, language, but also interplay constructively in some multisensorial areas of the brain.
% Impact
% Currently limited to perception and comprehension, this approach paves the way towards building an integrative model of the representations of the human brain. 
% which could unlock faithful in silico experimentation.
%   TRIBE holds the first place of the algonauts 2025 challenge.
\end{abstract}

\begin{figure}[htb]
    \centering
    \includegraphics[width=.6\linewidth]{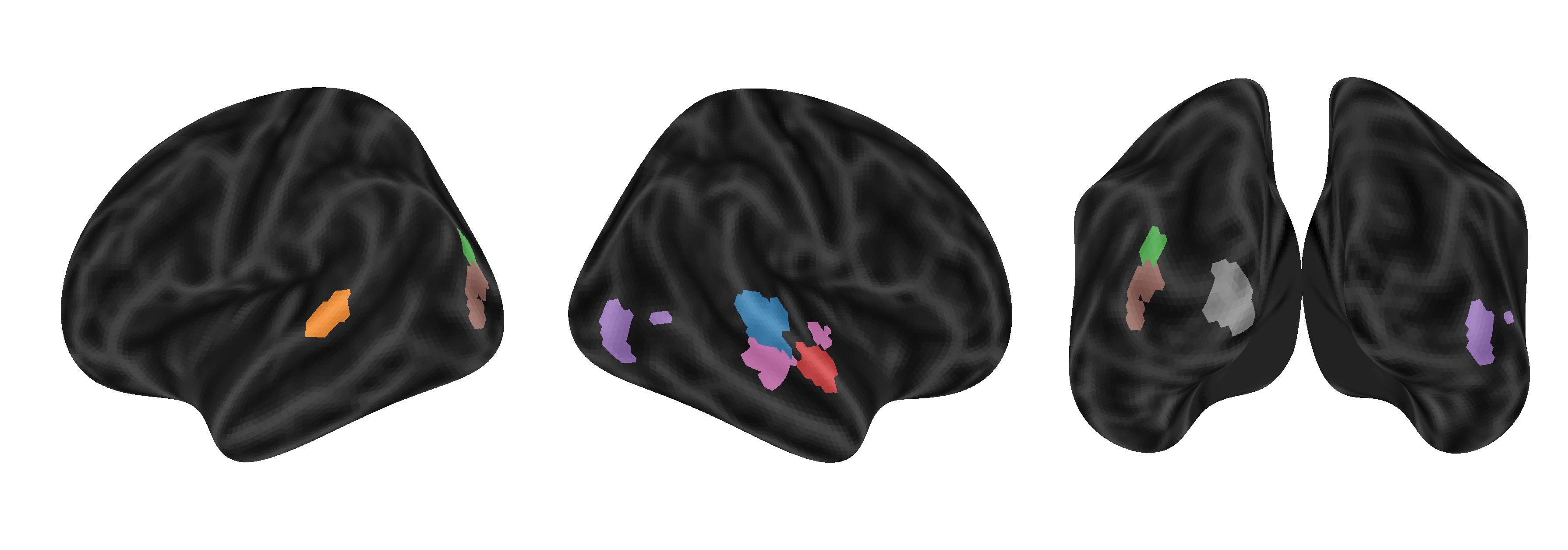}
    \includegraphics[width=\linewidth]{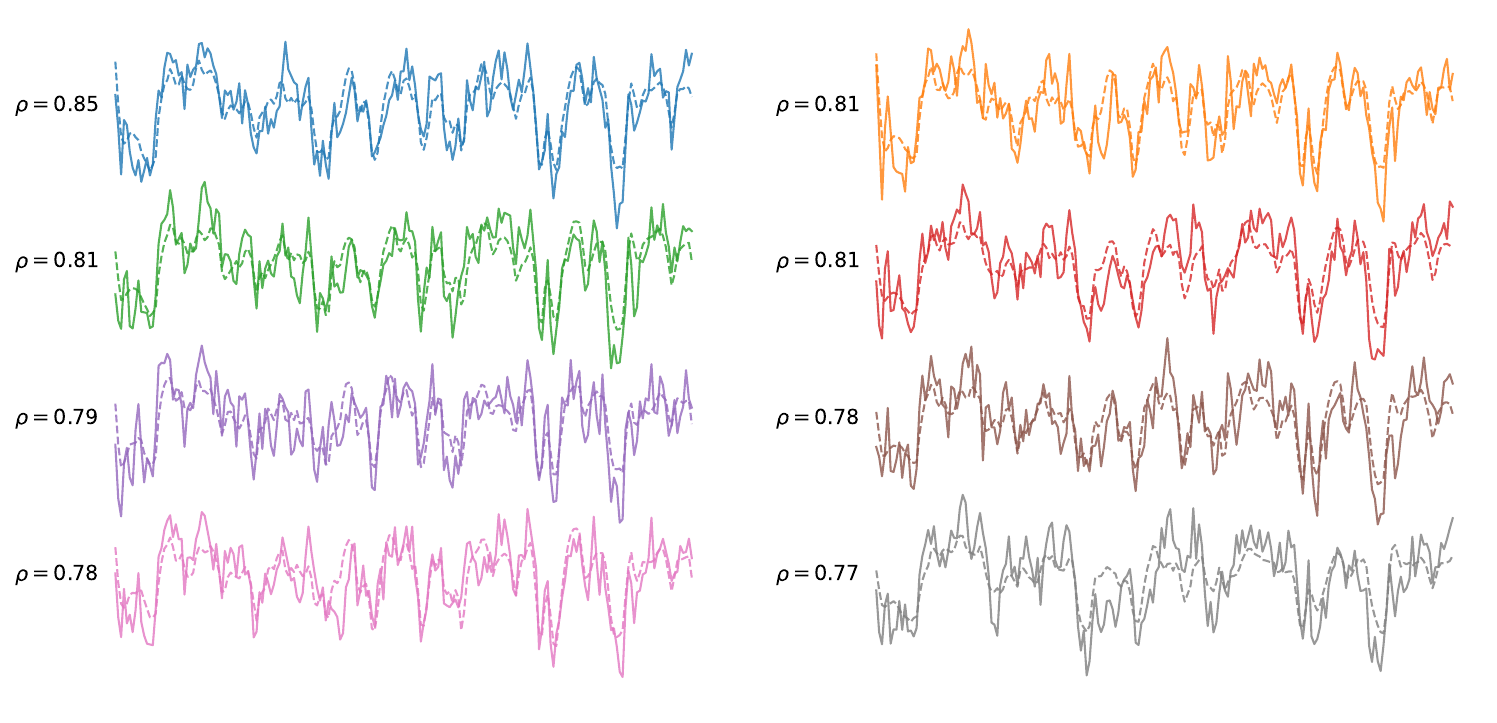}
    \caption{\textbf{TRIBE predicts brain responses to videos across diverse regions.} For eight brain parcels color-coded in the brain images, we report the BOLD response of the first participant to the first 5 minutes of a held-out movie in solid lines and our model's predictions in dashed lines, with the Pearson correlation of the two curves reported on the left.
    }
    \label{fig:demo}
\end{figure}

\section{Introduction}

\paragraph{Motivation.}

Progress in neuroscience has historically derived from an increasing specialization into cognitive tasks and brain areas. In the domain of vision, for instance, research focused on specialized cortical areas and their associated tasks, such as motion perception in V5~\cite{shadlen2001neural}, face recognition in the fusiform gyrus~\cite{kanwisher2006fusiform}, or the visual processing of written language in the visual word form area~\cite{dehaene2011unique}. While this divide-and-conquer approach has undeniably yielded deep insights into the brain's mechanisms of cognition, it has led to a fragmented scientific landscape: How neuronal assemblies together construct and globally broadcast a unified representation of the perceived world remains limited to coarse conceptual models~\cite{mashour2020conscious}.

The fast progress in AI in the domains of language~\cite{brown2020language,grattafiori2024llama}, image~\cite{oquab2023dinov2}, audio~\cite{baevski2020wav2vec,chung2021w2v} and video~\cite{tong2022videomae,assran2025v} may help resolve this fragmentation challenge. Indeed, the representations learnt by these AI models have been shown to -- at least partially -- align with those of the brain~\cite{yamins2014performance,huth2016natural,caucheteux2022brains, schrimpf2018brain}. 
Motivated by this unexpected alignment, several teams have built encoding models to predict brain responses to natural stimuli from the activations of neural networks in response to images~\cite{yamins2014performance}, speech~\cite{millet2022toward} and text~\cite{huth2016natural,caucheteux2022brains,toneva2019interpreting}. However, these encoding models are currently limited in three critical ways. 

First, \textit{linearity}: existing encoding approaches typically rely on a ridge regression to map the AI model representations onto those of the brain. This assumes that these two sets of representations are linearly equivalent -- a likely incorrect assumption~\cite{yamins2014performance, caucheteux2022brains,linsley2025can}. 

Second, \textit{subject-specificity}: due to large variability in brain responses from one subject to another, existing encoding approaches typically train a separate model for each subject, which prevents them from leveraging the similarities between brains (although~\cite{chehab2021deep}).

Third, \textit{unimodality}: most existing encoding approaches predict brain responses from unimodal stimuli, which makes them incapable of capturing how the brain integrates information from multiple modalities~\cite{hu2025neural}. This is particularly limiting as it has been shown that cross-modal interactions occur not only in specific multisensory areas~\cite{gao2023audiovisual,beauchamp2005see}, but also in primary sensory areas~\cite{driver2008multisensory,stein2008multisensory}.

\paragraph{Contribution}

In this work, we introduce TRIBE, a novel deep learning pipeline to predict the fMRI brain responses of participants watching videos from the corresponding images, audio and transcript. This approach addresses the three limitations outlined above: our model learns how to capture the dynamical integration of modalities in an end-to-end manner across the whole brain, and from multiple subjects. 

Our model achieves state-of-the-art results, reaching the first place out of 263 teams in the Algonauts 2025 competition on multimodal brain encoding. We demonstrate with ablation analyses the importance of the multimodal, multisubject and nonlinear nature of TRIBE. Finally, we observe that the benefit of multimodality is highest in associative cortices. %, where our model yields novel insights into multisensory integration.

\paragraph{Related work}

While there has been recent research on deep learning for multimodal brain decoding~\cite{dahan2025sim,xia2024umbrae}, there currently exists no equivalent for brain encoding. Some recent works suggest to train recurrent models to predict brain responses from frozen visual or linguistic features~\cite{gucclu2017modeling, chehab2021deep}, or fine-tune existing pretrained models using the brain encoding objective. While these relax the linearity assumption, they are restricted to a single sensory modality.

Conversely, a few recent studies have built encoding models on top of vision-language transformers, demonstrating gains compared to unimodal transformers~\cite{dong2023vision,oota2022visio,doerig2022semantic,wang2022incorporating,tang2023brain}. However, these works rely solely on linear mappings to model brain responses from the activations of the multimodal transformers. We believe this can be suboptimal for two reasons. First, multimodal transformers are still relatively new: at rare exceptions~\cite{jaegle2021perceiver,srivastava2024omnivec,abdin2024phi4}, they often only integrate static images and text (audio and video being significantly more compute-intensive), and tend to lag behind the performance of unimodal transformers. Second, and more fundamentally, the way these models integrate information across modalities may be very different from how the human brain does such multimodal integration. An ideal encoding pipeline should thus \textit{learn} how to best combine different modalities.

% \paragraph{Approach summary.}
% To address these challenges, we introduce TRIBE, the first deep encoding model trained jointly across subjects. 
\begin{figure}[htb]
    \centering
    \includegraphics[width=\linewidth]{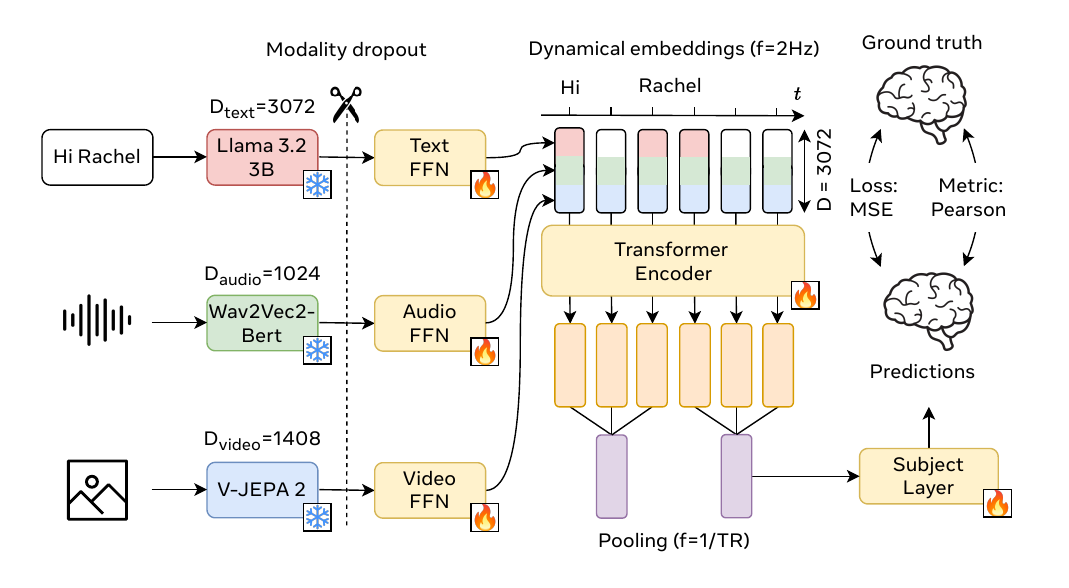}
    \caption{\textbf{Visual summary of our method.}}
    \label{fig:sketch}
\end{figure}

\section{Methods}
\label{sec:methods}

\subsection{Overview}

\paragraph{Task}

Our objective is to predict the brain activity of participants watching videos. This is framed as a regression task where the targets are the blood-oxygen-level-dependent (BOLD) signals detected at every repetition time (TR=1.49s) of a 3T fMRI recording device, separated into 1,000 non-overlapping parcels \cref{fig:demo}.

For this, we take as input the video clip being viewed by the participant, as well as the corresponding audio file and transcript. From these, we extract high-dimensional embeddings from the intermediate layers of start-of-the art generative AI models along three modalities of interest: text, audio and video, which we feed to our deep encoding model, as illustrated in \cref{fig:sketch}.

\paragraph{Evaluation}
To assess performance, we evaluate for each parcel the Pearson Correlation $\rho$ between predicted and ground truth fMRI responses across all TRs of the evaluation set. We then average these values over all parcels. We also refer to this metric as the ``encoding score'' of the model.

We hold out 10\% of the recording sessions for the validation set,  ensuring that the same videos are held out for each participant to prevent any form of data leakage.

% We also assess how well the predictions of TRIBE can be used to perform stimuli retrieval, where the objective is to find which video clip out of a retrieval set was viewed by a participant, given its BOLD response. For each clip in the retrieval set, we compute the predicted BOLD responses with TRIBE, and pick the one whose prediction is closest to the true BOLD response.

\subsection{Dataset}

\paragraph{Data collection}
We train our encoding model on the Courtois NeuroMod dataset~\cite{st2023cneuromod}. This dataset consists of six human participants who watched the same naturalistic videos, namely the first six seasons of the popular TV series \textit{Friends} as well as four movies: \textit{The Bourne Supremacy, Hidden Figures, The Wolf of Wall Street} and \textit{Life} (a BBC Nature documentary). This amounts to an unprecedently large recording volume of over 80 hours of fMRI per subject. In the present work, we focus on a subset of four subjects curated for the Algonauts 2025 competition~\cite{gifford2024algonauts}.

\paragraph{Preprocessing}
We use the preprocessing pipeline provided by the Algonauts2025 competition. The whole-brain BOLD fMRI responses are preprocessed using fMRIprep~\cite{esteban2019fmriprep} and projected to the Montreal Neurological Institute
(MNI152NLin2009cAsym) standard space ~\cite{brett2002mni}. Functional images are then parcellated by averaging voxel-wise BOLD signals within each of the 1,000 parcels of the Schaefer atlas~\cite{schaefer2018local}, yielding a single fMRI time series for each of the parcels. Finally, activations were z-scored per parcel across each session (corresponding to approximately 15 minutes of recording).

% \paragraph{Training and validation splitting}
% Data leakage is known to be a severe issue when training encoding or decoding models~\cite{jo2024eeg}, given that stimuli are typically repeated across participants in neuroscientific experiments. To mitigate this, we consider two settings: (i) in the \textit{in-distribution} setting, we randomly select 10\% of the \textit{sessions} to hold, keeping the same splits across participants; (ii) in the \textit{out-of-distribution (OOD)} setting, we randomly hold out entire \textit{movies}.

% \paragraph{Lebel}
% The dataset collected by \citet{lebel2023natural} comprises 7 subjects exposed to 6 hours of podcasts each. This dataset does not contain any visual stimuli, and therefore serves as a complementary testbed for TRIBE.

\subsection{Model}

% In this section, we describe our encoding pipeline: please refer to \cref{fig:sketch} for a visual summary. 

\paragraph{Timed text embeddings}
We extract "timed" text embeddings from the timestamped transcripts of the videos. For each word $w$ to embed, we prepend the preceding $k=1,024$ words in the transcript, which we feed through Llama-3.2-3B~\cite{grattafiori2024llama}. For each intermediate layer $l$, we extract the token(s) overlapping with the word $w$ and average them to obtain a contextualized word embedding of dimension $D_\text{text}=2048$.

We then construct an evenly spaced grid at a frequency $f=2$\,Hz, and for each time-bin $\tau$, we sum the embeddings of words which overlap with the bin. 
% Note that this method for obtaining dynamical contextual text embeddings is, to the best of our knowledge, a novel technical contribution. %?? JR No this is what we've been doing with charlotte for many years?
% It is an important component of our pipeline as it 
This allows to temporally align the text features with the audio and video features. 
% Formally, the timed text embeddings of layer $l$ are given by:
% \begin{equation}
%     \mathbf {TTE}^l(\tau) = \sum_{w\in \Omega(\tau)} \mathbf {V}^l_w, \quad \Omega(\tau)=\left\{ w \ |\  w \ \cap\  \tau \neq  \emptyset \right\}
% \end{equation}

\paragraph{Audio embeddings}
To obtain audio embeddings, we extract audio files from the videos, split them into 60-second chunks, then feed these through Wav2Vec-Bert-2.0~\cite{chung2021w2v}. We then resample the hidden representations of the latter from 50\,Hz to $f=2$\,Hz. For each intermediate layer $l$, this yields time series of embeddings of dimension $D_\text{audio}=1,024$.

Note that the resulting embeddings carry bidirectional information about both the past and future of the stimulus window, whereas text and video embeddings only contain information about the past.

\paragraph{Video embeddings}
For video embeddings, we again construct an evenly spaced grid at a frequency $f=2$\,Hz, and for each bin of time, we feed 64 frames spanning the preceding 4 seconds to Video-JEPA 2 gigantic~\cite{assran2025v}. For each intermediate layer $l$, we compress the tensor of activations by averaging over all patch tokens, yielding a time series of embeddings of size $D_\text{video}=1,280$.

Note that this spatial averaging step was necessary to keep the size of the tensor manageable. However, it comes at the cost of discarding positional information, which we expect to deteriorate encoding performance in low-level visual areas which exhibit a retinotopic mapping~\cite{wandell2011imaging}.

\paragraph{Combining the modalities}

For each of the three modalities $m$, the feature extraction described above leads to a time series of embeddings at $f=2$\,Hz, with embeddings of shape $[L_m,D_m]$, where $L_m$ and $D_m$ are the number of layers and dimensionality of the transformer of modality $m$. 

To compress these embeddings while retaining both low-level and high-level information, for each modality, we split the layers into $L$ groups, then average the tensor per group along the layer dimension, compressing to a shape $[L, D_m]$. 

We then concatenate the layers and feed the resulting vector through a linear layer with a shared output dimension $D=1024$ followed by layer normalization. Finally, we concatenate the three modalities, leading to a time series of \textit{multimodal} embeddings of shape $3\times 1024$. This will be the input to our transformer encoder.

% Not only does this enable TRIBE to be applied out-of-the box to different datasets, it also improves generalization in the original training task as shown in \cref{app:modality-dropout}.

\paragraph{Transformer encoder}

We extract windows of duration $T=N\times TR$ from these embedding time series, add learnable positional embeddings and a learnable subject embedding, then feed the result through a Transformer encoder with 8 layers and 8 attention heads. This enables information to be exchanged between timesteps.

At the output of the transformer, we use an adaptive average pooling layer to compress the sequence from length $fT$ to $N$, yielding one embedding per TR. Following~\cite{defossez2023decoding}, we then use a subject-conditional linear layer to project the latter to the 1,000-dimensional target space. 

\subsection{Training details}

\paragraph{Modality dropout}

One desirable property of a multimodal encoding model is its ability to provide meaningful predictions in the absence of one or several modalities, for example for a silent movie or a podcast. To encourage this behaviour, while at the same time avoiding excessive reliance on one modality, we introduce modality dropout: during training, we randomly mask off each modality by zeroing out the corresponding input tensor with a probability $p$, ensuring that at least one modality is left unmasked.

\paragraph{Optimization}

We train our model for up to 15 epochs with the AdamW optimizer~\cite{loshchilov2017decoupled} using a batch size of 16. The learning rate is warmed up linearly to $10^{-4}$ over the first 10\% of steps, then decayed following a cosine learning rate schedule. We use early stopping based on the validation Pearson score computed on a held-out set.
To improve generalization, we use stochastic weight averaging~\cite{izmailov2018averaging}, which involves averaging model weights obtained at the end of each epoch, once the validation metrics are near their plateau.

\paragraph{Ensembling}

To further improve generalization, we ensemble the predictions of $M=1000$ models, whose initialization and shuffling seeds are all different. To strengthen ensemble diversity, for each model, we sample a set of hyperparameters uniformly in the grid specified in \cref{app:hyperparameters}. For each parcel separately, we compute the encoding score of all models on the validation set, then compute a softmax distribution over models with temperature $0.3$, which will determine the weight assigned to each model for this given parcel.

\paragraph{Implementation details}

We extract stimuli features from pretrained language, audio and video models available on the \texttt{HuggingFace} platform~\cite{jain2022hugging} and cache them as \texttt{Numpy} memmap arrays~\cite{harris2020array} for fast loading during the training of our encoding model. Feature extraction is completed in 24 hours on 128 V100 GPUs with 32GB of VRAM, and model training lasts 24 hours on a single such GPU. We use the transformer implementation from the \texttt{x-transformers} package\footnote{\url{https://github.com/lucidrains/x-transformers}}. We list the licenses of the assets used in this work in \cref{app:licenses}.

\begin{table}[htb]
\centering
\begin{tabular}{clccccc}
\toprule
\textbf{Rank} & \textbf{Team} & \textbf{Mean score} & \textbf{Subject 1} & \textbf{Subject 2} & \textbf{Subject 3} & \textbf{Subject 5} \\ 
\midrule
1 & \textbf{Ours}    & \textbf{0.2146} & 0.2381 & 0.2105 & 0.2377 & 0.1720 \\
2 & NCG         & \textbf{0.2096} & 0.2353 & 0.2046 & 0.2268 & 0.1718 \\
3 & SDA         & \textbf{0.2094} & 0.2233 & 0.2072 & 0.2271 & 0.1798 \\
4 & MedARC      & \textbf{0.2085} & 0.2295 & 0.2003 & 0.2300 & 0.1743 \\
5 & CVIU-UARK   & \textbf{0.2055} & 0.2306 & 0.2010 & 0.2240 & 0.1662 \\
\bottomrule
\end{tabular}
\vspace{0.1cm}
\caption{\textbf{Our model achieves first place in the Algonauts 2025 public leaderboard.} We report the results of the top five out of 263 teams.}
\label{tab:leaderboard}
\end{table}

\begin{table}[htb]
\centering
\begin{tabular}{clccccc}
\toprule
\textbf{OOD} & \textbf{Movie} & \textbf{Mean score} & \textbf{Subject 1} & \textbf{Subject 2} & \textbf{Subject 3} & \textbf{Subject 5} \\ 
\midrule
\xmark & Friends Season 7 & \textbf{0.3195} & 0.3419 & 0.3239 & 0.3346 &	0.2775 \\
\midrule
\cmark & Pulp Fiction & \textbf{0.2604} & 0.2765 & 0.2611 & 0.2431 & 0.2610 \\
\cmark & Princess Mononoke & \textbf{0.2449} & 0.2816 & 0.2507 & 0.2851 & 0.1623 \\
\cmark & Passe-partout & \textbf{0.2323} & 0.2763 & 0.2587 & 0.2370 & 0.1573 \\
\cmark & World of Tomorrow & \textbf{0.1924} & 0.2210 & 0.1606 & 0.2196 & 0.1686 \\
\cmark & Planet Earth & \textbf{0.1886} & 0.1483 & 0.2029 & 0.2331 & 0.1699 \\
\cmark & Charlie Chaplin & \textbf{0.1686} & 0.2249 & 0.1289 & 0.2080 & 0.1128 \\
\bottomrule
\end{tabular}
\vspace{0.1cm}
\caption{\textbf{Our model generalizes to highly out-of-distribution movies.} We provide the detailed results on the held-out datasets of the Algonauts 2025 competition.}
\label{tab:leaderboard-ood}
\end{table}

\section{Results}

\subsection{Algonauts 2025 competition results}

Our model achieves the first place out of 262 teams in the Algonauts 2025 competition on multimodal brain encoding. As shown in \cref{tab:leaderboard}, we outperform competitors by a substantial margin: the gap between our model and the runner-up is larger than between the runner-up and the fifth.

We display the results across the various held-out datasets in \cref{tab:leaderboard-ood}. In the in-distribution setting of the first phase of the competition (\textit{Friends} season 7), our model achieves a mean score of 0.3195. Unsurprisingly, when tested in the out-of-distribution setting of the second phase of the competition, performance is lower, with an average of 0.2146. Remarkably, our model achieves robust scores even in the extreme out-of-distribution setting of cartoons (0.1924 for \textit{World of Tomorrow}) wildlife documentaries (0.1886 for \textit{Planet Earth}), and silent black-and-white movies (0.1686 for \textit{Charlie Chaplin}). 

\begin{figure}[htb]
    \begin{subfigure}[b]{.34\linewidth}
        \centering
        \includegraphics[width=\linewidth]{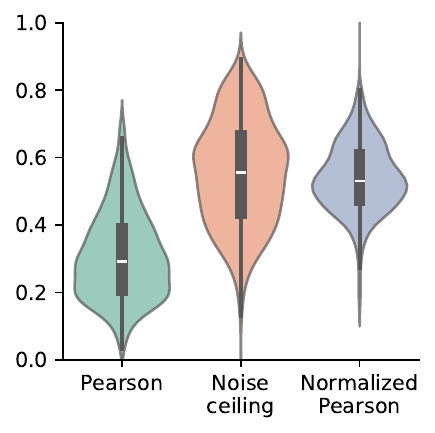}    
        \caption{Distribution across parcels}
    \end{subfigure}
    \begin{subfigure}[b]{.65\linewidth} 
        \centering
        \includegraphics[width=\linewidth]   {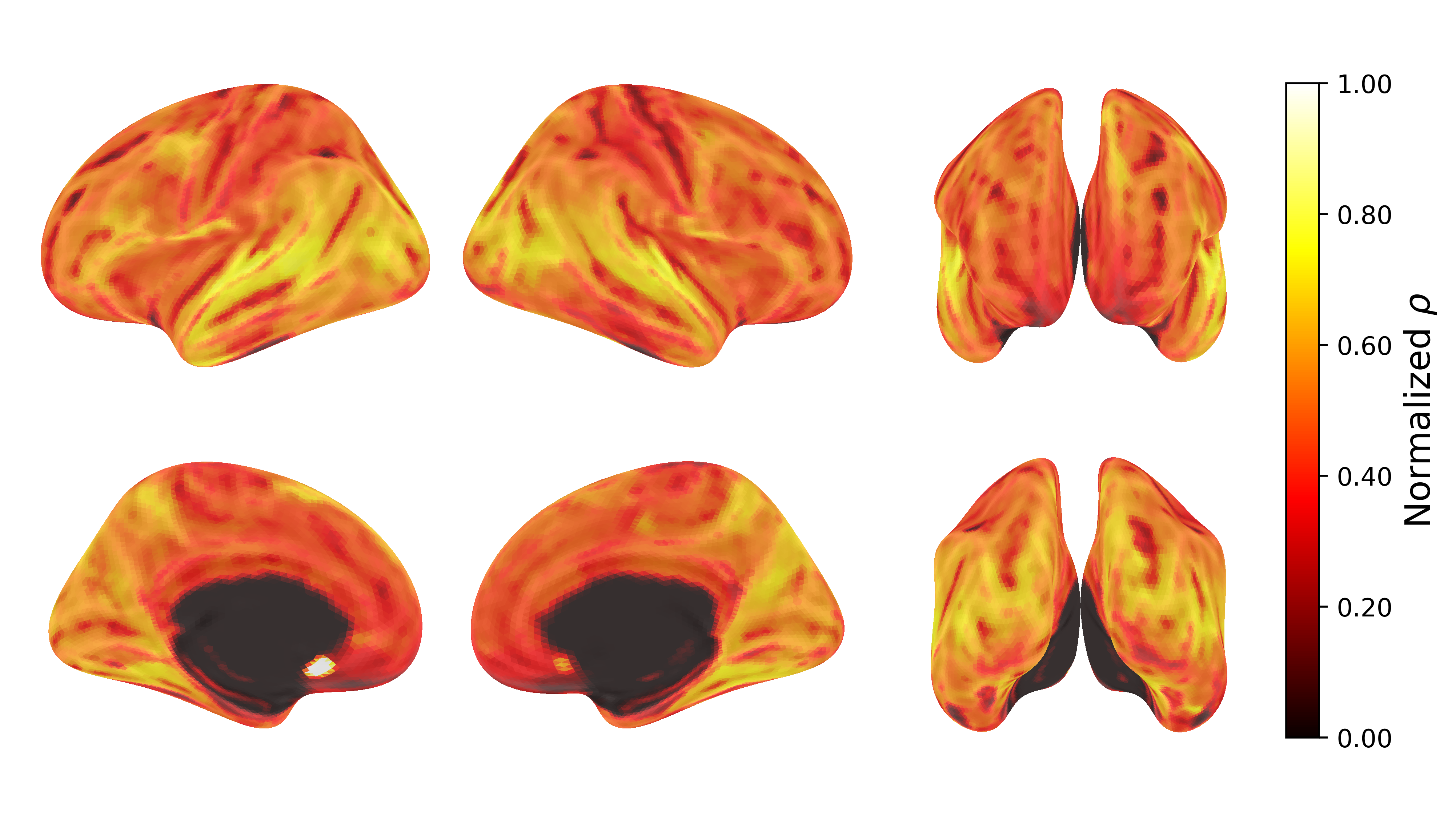}
        \caption{Cortical projection of the normalized scores}
    \end{subfigure}
    \caption{\textbf{Our model performs consistently across the whole brain.}\\
    (a) We report a histogram of unnormalized and normalized encoding scores (see definition in \cref{eq:noise_ceiling}) of the 1000 parcels, averaged across subjects.\\
    (b) We project the normalized encoding scores of the first subject onto the $\texttt{fsaverage5}$ cortical surface. \\
    }
    \label{fig:noise-ceiling}
\end{figure}

\subsection{Noise ceiling analysis}

Following common practice in the brain encoding literature~\cite{schoppe2016measuring}, we estimate the noise ceiling to investigate to what extent encoding errors are due to unexplainable randomness rather than model suboptimality. We achieve this by computing the inter-trial correlation $\rho_{self}$ of the BOLD responses to the movies \textit{Hidden Figures} and \textit{Life}, which were viewed twice by each participant. We then define the normalized Pearson correlation of our model by dividing it by that of an ideal encoding model, following~\cite{schoppe2016measuring}:
\begin{equation}
    \rho_{norm} = \frac{\rho}{\rho_{max}}, \quad \rho_{max} = \sqrt{\frac{2}{1+\frac{1}{\rho_{self}}}}
    \label{eq:noise_ceiling}
\end{equation}

Our model achieves a normalized Pearson correlation of $0.54\pm 0.1$ across all parcels (\cref{fig:noise-ceiling}a): in other words, it captures 54\% of explainable variance on average. The fairly small inter-quartile range reflects the fact that our model is rather consistent across brain areas. The highest values are achieved in the auditory and language processing cortices, where our model is near the noise ceiling (\cref{fig:noise-ceiling}b).

\begin{figure}[htb]
    \centering
    \begin{subfigure}[b]{.34\linewidth}
        \centering
        \includegraphics[width=\linewidth]{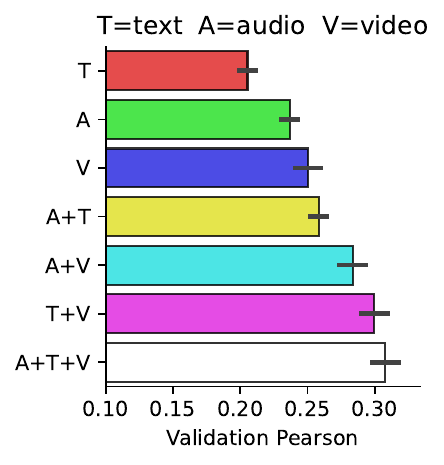}
        \caption{Modality ablations}
    \end{subfigure}
    \begin{subfigure}[b]{.65\linewidth}
        \includegraphics[width=\linewidth]{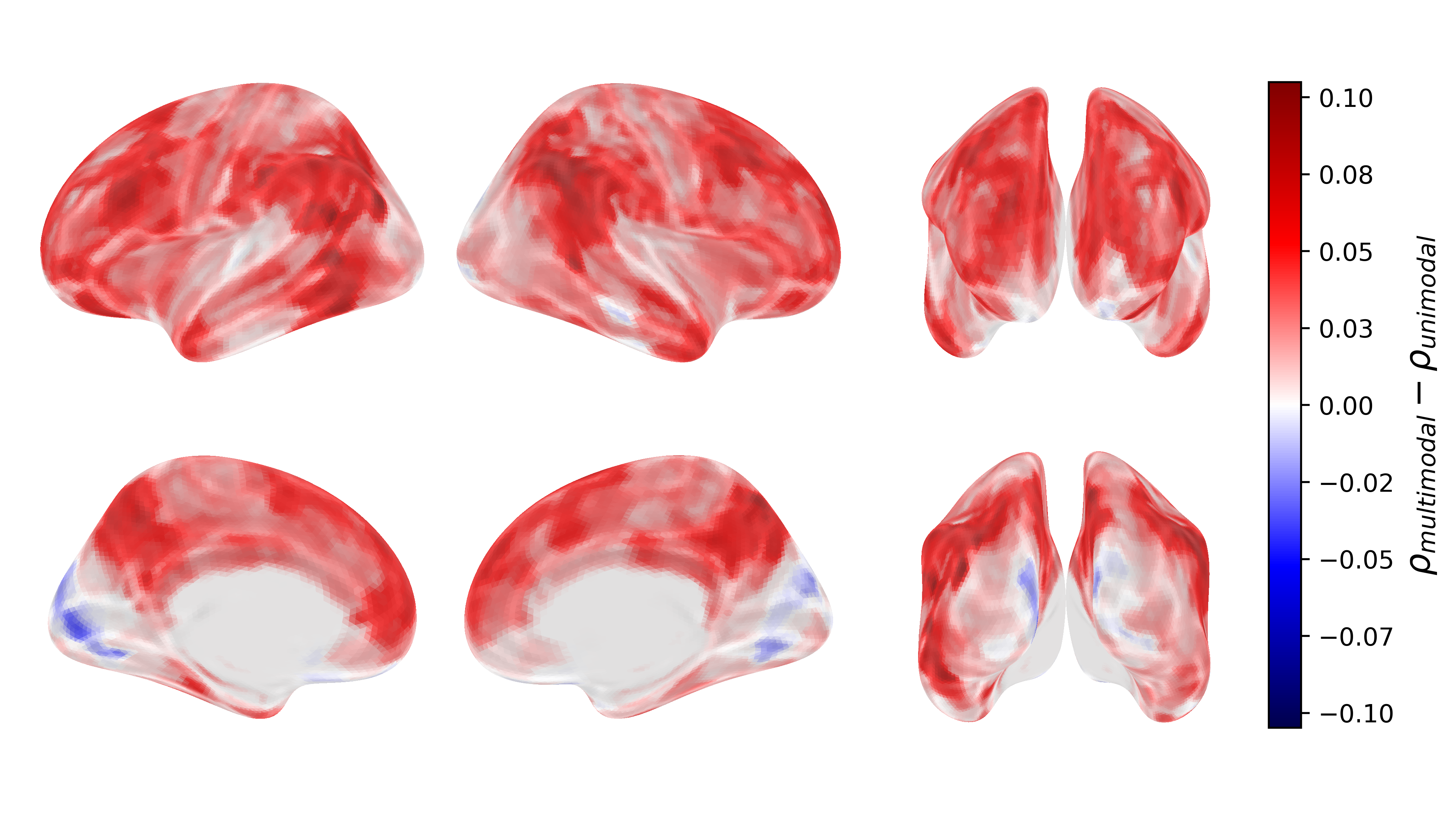}
        \caption{Multimodal versus best unimodal}
    \end{subfigure}
    \caption{
    \textbf{Multimodal encoders outperform unimodal encoders.}\\
    (a) We compare the encoding scores of encoders trained on a subset of modalities in the same conditions. Error bars denote s.e.m across subjects.\\
    (b) For each parcel of the first subject, we compute the difference in encoding score between the multimodal encoder and the best out of the three unimodal encoders, then project the results onto the \texttt{fsaverage5} cortical surface.
    }
    \label{fig:multimodal-vs-unimodal}
\end{figure}
\begin{figure}[htb]
    \centering
    \begin{subfigure}[b]{.49\linewidth}
        \centering
        \includegraphics[width=.4\linewidth]{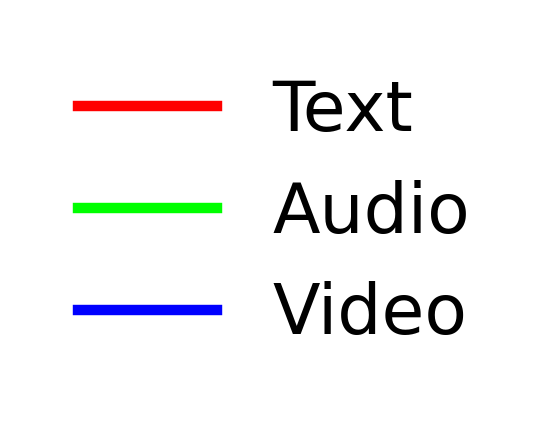}        
        \includegraphics[width=\linewidth]{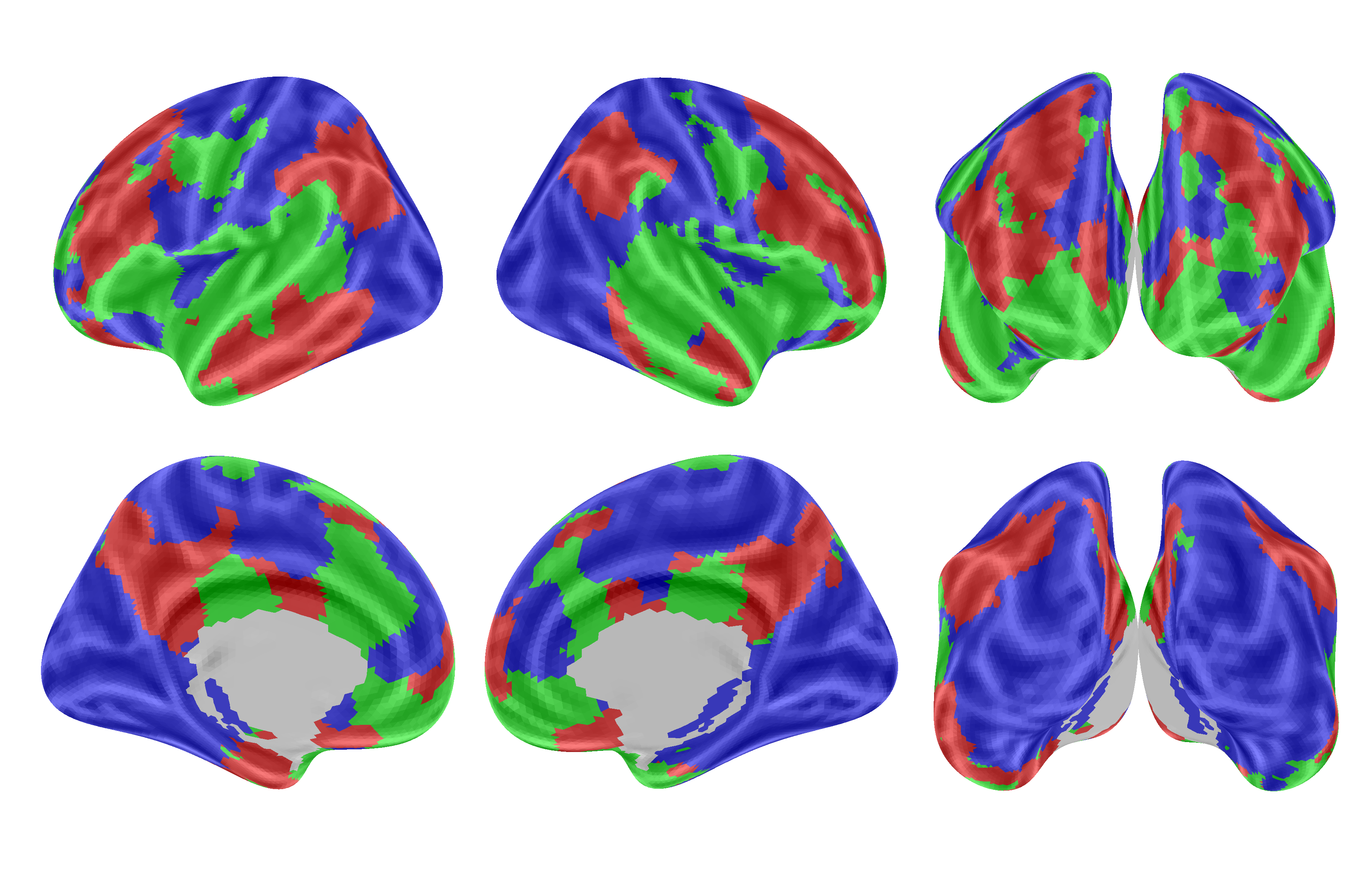}
        \caption{Highest scoring modalities}
    \end{subfigure}
    \hfill
    \begin{subfigure}[b]{.49\linewidth}
        \centering
        \includegraphics[width=.7\linewidth]{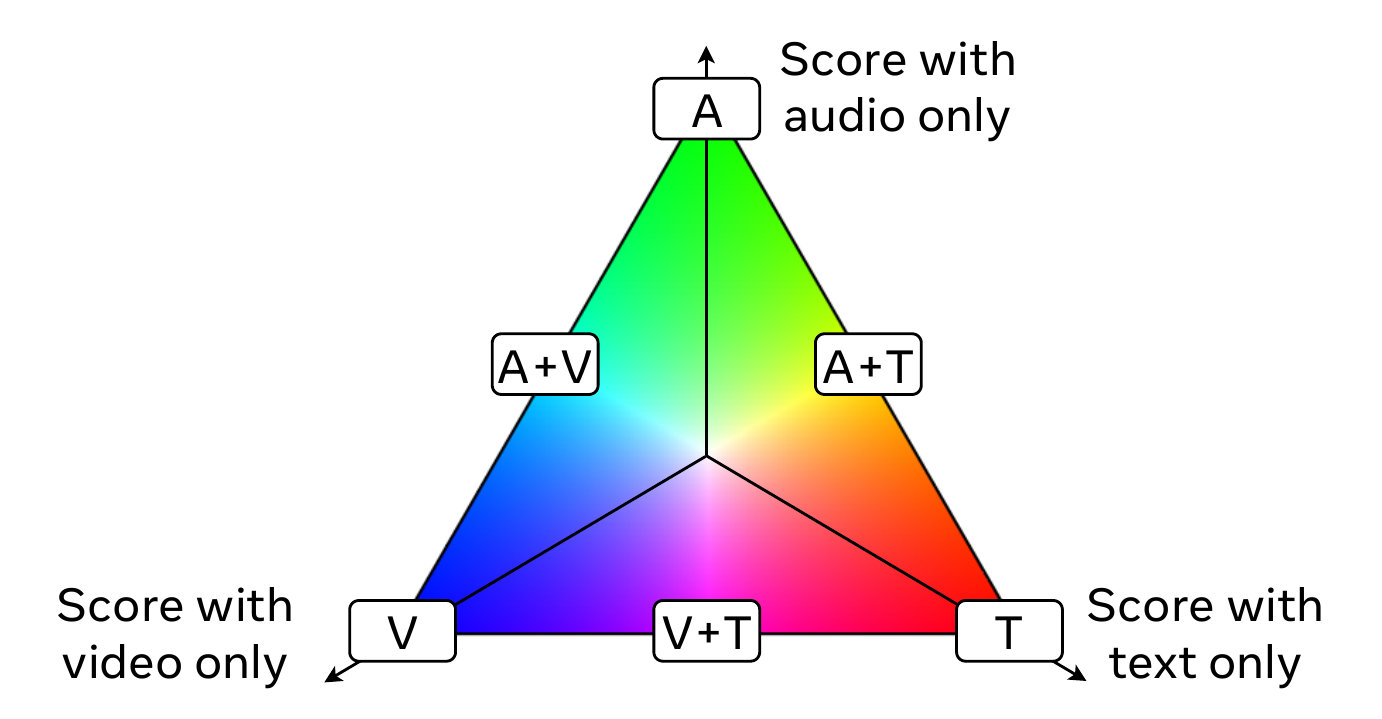}
        \includegraphics[width=\linewidth]{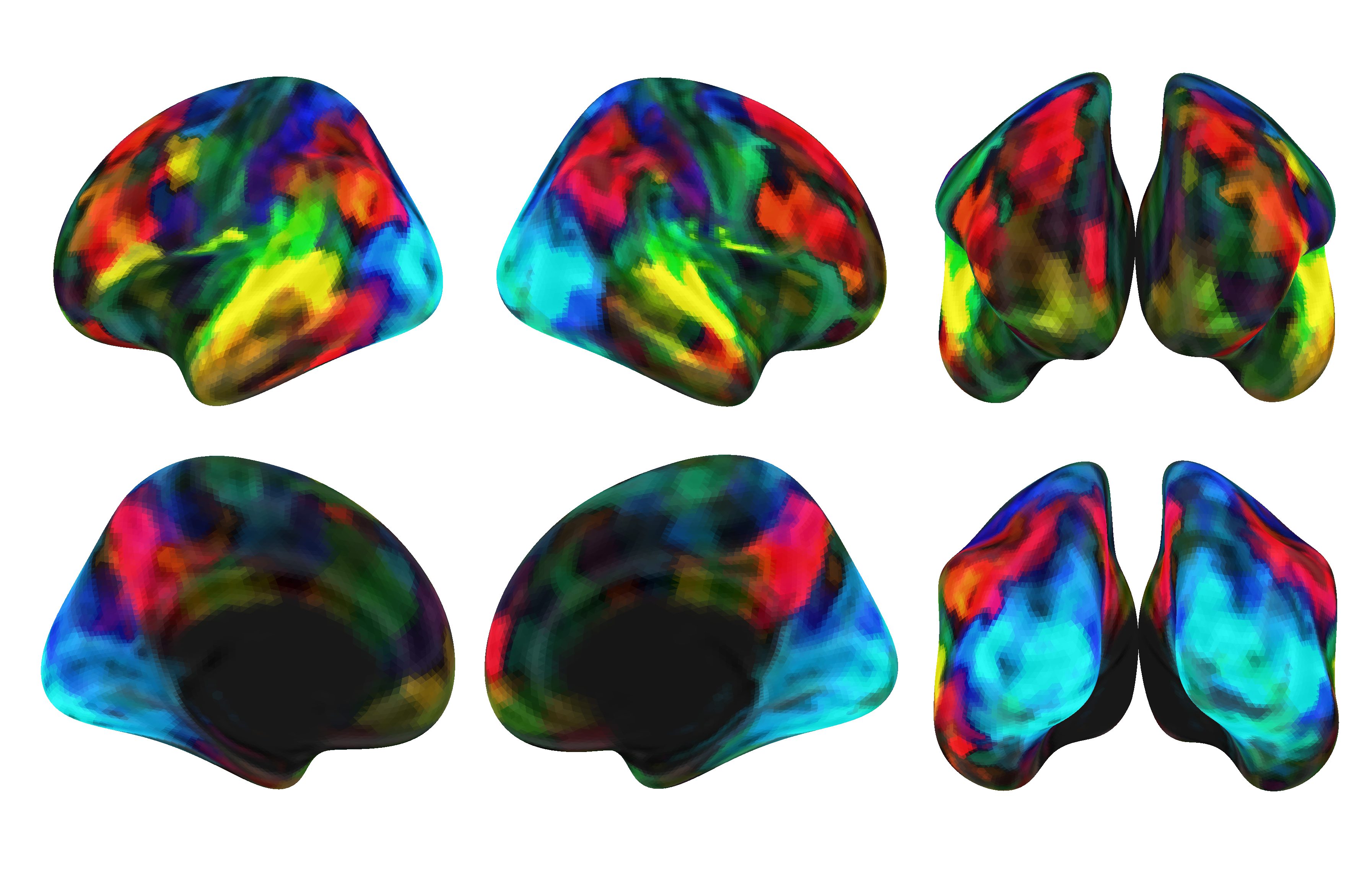}
        \caption{Interplay between modalities}
    \end{subfigure}
        \caption{
    \textbf{The modalities and their interplay map onto the expected brain areas.}\\
    (a) For each parcel of the first subject, we assess the encoding contribution of each modality by masking the two other modalities, and color-code according to the highest contribution. \\
    (b) We color-code each parcel using an RGB encoding where red, green and blue intensities are determined by the encoding score of the model with text, audio and video unmasked, respectively. For readability, we limit ourselves to bimodal interactions by substracting the smallest of the three contributions. \textcolor{red}{Red}, \textcolor{blue}{blue} and \textcolor{green}{green} areas correspond to unimodal areas well encoded by text, audio and video respectively, while \textcolor{magenta}{magenta}, \textcolor{yellow}{yellow} and \textcolor{cyan}{cyan} correspond to bimodal areas well encoded by text-video, text-audio and video-audio interactions respectively.
    }
    \label{fig:modalities}
\end{figure}
\begin{figure}[htb]
    \centering
    \begin{subfigure}[b]{.32\linewidth}
    \includegraphics[width=\linewidth]{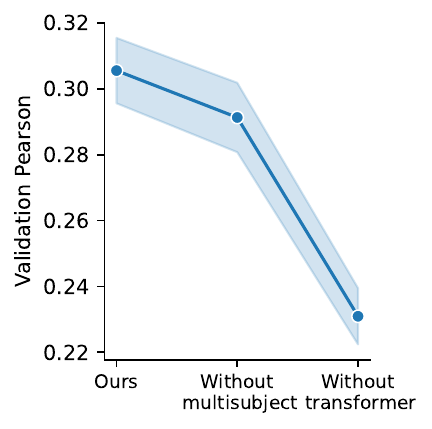}
    \caption{Model ablations}
    \end{subfigure}
    \begin{subfigure}[b]{.32\linewidth}
    \includegraphics[width=\linewidth]{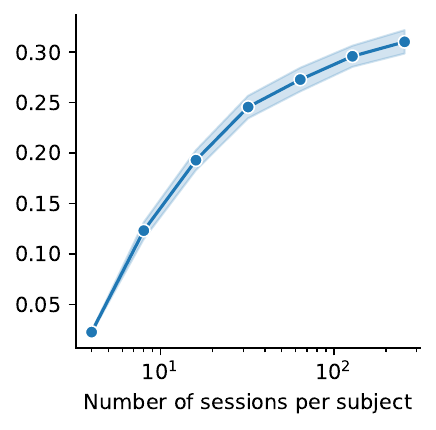}
    \caption{Number of sessions}
    \end{subfigure}
    \begin{subfigure}[b]{.32\linewidth}
    \centering
    \includegraphics[width=\linewidth]{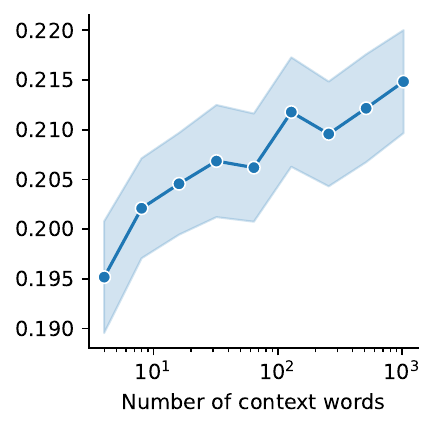}
    \caption{Context length}
    \end{subfigure}
    \caption{
    \textbf{Ablations and scaling laws of our model.}\\
    (a) We compare the results of our model with those achieved without multi-subject training or without the transformer model.\\
    (b) We report the results obtained by the multimodal encoder as we increase the number of sessions used in the training set.\\
    % (b) We report the validation encoding score obtained by the text-only encoder as we increase the size of the pretrained language model used to extract features, for a context length of 128 words.\\
    (c) We report the results obtained by the text-only encoder as we increase the context length of the language model used to extract features (Llama-3.2-3B).\\
    In all panels, shaded regions indicate the s.e.m over the four subjects.
    }
    \label{fig:scaling}
\end{figure}

% The encoding abilities of TRIBE are illustrated by the leaderboard of Algonauts 2025 competition, shown in \cref{tab:leaderboard}: as of 25th of April, our model achieves the first place out of 160 participants (note that this competition is ongoing, hence the leaderboard may vary).

% We assess the in-distribution performance of our model in \cref{fig:metrics}a by testing it on a held-out season of Friends. Our model achieves a grand-averaged Pearson correlation of 0.31, with a rather small spread across participants: 0.32 for the best and 0.28 for the worst. We also assess out-of-distribution performance by retraining models with entire movies held out, and observe that our model is robust in this setting, with subject-averaged Pearson scores ranging from 0.25 for the best movie (the Wolf of Wall Street) to 0.19 for the worst (Life, a wildlife documentary which is particularly out-of-distribution). 

\subsection{The benefit of multimodality}

To what extent do the three modalities combined by TRIBE contribute to encoding performance? We address this question in \cref{fig:multimodal-vs-unimodal}a, by assessing the encoding performance of TRIBE retrained with various modalities ablated. When training on a single modality, TRIBE achieves substantially lower encoding scores. We observe that the text modality achieves the lowest average encoding score with 0.22, followed by audio at 0.24 and video at 0.25. When combining any two modalities together, the encoding scores rise significantly compared to the unimodal models: the best bimodal model, obtained by combining text and video, achieves an encoding score of 0.30. Finally, combining the three modalities together yields an additional boost, bringing the value to 0.31. This hints to the fact that each modality plays a complementary role.

In which areas does multimodality yield the strongest gains? In \cref{fig:multimodal-vs-unimodal}b, we compare for each parcel the encoding score of the multimodal encoder with that of the best of the three unimodal encoders. We observe that the multimodal encoder consistently outperforms the unimodal models, especially in associative areas such as the prefrontal or parieto-occipito-temporal cortices (up to 30\% increase in encoding score). Interestingly, the multimodal model performs less well than the vision-only model in the primary visual cortex, which is highly specific to visual features. 

Together, these results demonstrate that our multimodal encoder effectively captures interactions between modalities, which improves whole-brain decoding.

\subsection{The interplay between modalities}

Which brain areas is dominated by unimodal or multimodal representations? To address this question, we single out modalities by probing our multimodal model with all modalities masked off except one (using the same procedure as for the dropout described in \cref{sec:methods}).

In \cref{fig:modalities}a, we color each parcel of the cortex according to the modality that achieves the highest encoding score via this procedure. The three modalities each cover broad regions: audio predominates near the temporal gyrus, video predominates in the occipital cortex and parts of the parietal cortex, while the text feature, which presumably contains the most semantic information, predominates in large parts of the parietal and prefrontal lobes.

To study the interplay between modalities, we then overlay the contribution of the three modalities using an RGB encoding where red, green and blue respectively represent the encoding scores achieved solely with text, audio and video (\cref{fig:modalities}b). To make hues more visible, we restrict our analyses to unimodal and bimodal interactions by substracting the smallest contribution among the three modalities. We observe interesting bimodal associations in some key areas: in particular, %text+video (magenta) is observed some motor and sensory association cortices, % JR: not very cleary
text+audio (yellow) in can be observed in the superior temporal lobe and video+audio (cyan) can be observed in the ventral and dorsal visual cortices.

These observations provide new insights %not only align qualitatively with what one would expect from a neuroscientific point of view, but also yield valuable insights 
on how multisensory integration may occurs in the human cortex.

\subsection{Ablations and scaling laws}

In \cref{fig:scaling}a, we demonstrate the importance of using a transformer and a multi-subject training scheme: the encoding score drops from 0.31 to 0.29 when training separately for each subject, and down to 0.23 when removing the transformer.

In \cref{fig:scaling}b, we show how the encoding score scales with the amount of recordings in the training set. We observe a strong increasing trend which has not reached a plateau, in line with recent work~\cite{antonello2023scaling}.

In \cref{fig:scaling}c, we show that increasing the context length used for the language model words strongly enhances encoding performance, without any plateau even at very long context lengths of 1024 words. This confirms that TRIBE effectively captures high-level semantic features, far beyond the word and sentence level.

\section{Discussion}

In this work, we make a step towards an integrative model of the brain during naturalistic perception by training an encoding model on an unprecedently-large fMRI dataset of participants watching videos. Crucially, our model is the first encoding pipeline which is simultaneously nonlinear, multisubject and multimodal: our ablations demonstrate the importance of these three aspects for encoding, especially in associative cortices. Our model achieves the first place in the Algonauts 2025 brain encoding competition, and scaling laws suggest that encoding performance increases systematically with the number of recordings, holding promise for further improvements with larger datasets.

\paragraph{Limitations}
There are several remaining limitations to our work. First, our approach currently operates on a coarse parcellation of brain areas -- reducing hundreds of thousands of voxels to 1,000 cortical parcels. This design choice, introduced by the Algonauts2025 challenge, likely increases the signal-to-noise ratio by smoothing out the responses spatially, and certainly reduces computational cost, which is important for the whole-brain prediction setting considered here. However, this approach limits the spatial resolution of our model, which inherently prevents it from capturing highly localized phenomena. Adapting our model for voxel-level prediction is an important avenue for future work. Second, our current approach is limited to fMRI data. Consequently, the precise temporal dynamics of neuronal activity, and the exact neural assemblies underlying these macroscopic signals, remain, here, unresolved. Third, while the volume of recording per participant in the study considered here is particularly large, only four participants were included: extending and improving our results on a larger pool of participants is an important next step.
Finally, the present approach remains limited to perception and comprehension. Behavior, memory and decisions are other important cognitive components to integrate into the present model. 

\paragraph{Broader impact}
% Building accurate predictors of the brain paves the way towards a new realm in neuroscience: that of \textit{in silico} experimentation, where costly experiments involving repetitions across subjects could be complemented with simulated data using encoding models~\cite{jain2024computational}. We aim for our whole-brain encoding framework to chart a viable path toward a foundational model of cognition -- one that bridges the diverse avenues of neuroscientific research. 
Building a model able to accurately predict human brain responses to complex and naturalistic conditions is an important endeavour for neuroscience. Not only does this approach open the possibility of exploring cognitive abilities (e.g. theory of mind, humour) that are challenging to isolate with minimalist designs, but they will eventually be necessary to evaluate increasingly complex models of cognition and intelligence. In addition, our approach forges a path to (1) integrate the different sub-fields of neuroscience into a single framework and to (2) develop in silico experimentation~\cite{jain2024computational}, where in vivo experiments could be complemented and guided by the predictions of a brain encoder. While the exploration of this paradigm falls beyond the scope of this technical report, we believe that epistemology is ordered: interpretation and scientific insights are most legitimate if they derive from the model that makes the best prediction. In that regard, the first place achieved by TRIBE in the Algonauts 2025 competition gives scientific credit to our approach.

% We do however call for caution in the potential misuses of this tool, in particular in the field of neuromarketing, where predicting brain responses to advertizing could exploit our cognitive weaknesses to amplify harmful consumption habits~\cite{murphy2008neuroethics}.

%We hope that this work sparks more research in the former direction.

% \paragraph{Acknowledgements}
% The authors would like to thank the organizers of the Algonauts 2025 competition for making this work possible.

\clearpage

\bibliographystyle{unsrt}
\bibliography{refs}

\begin{thebibliography}{10}

\bibitem{shadlen2001neural}
Michael~N Shadlen and William~T Newsome.
\newblock Neural basis of a perceptual decision in the parietal cortex (area lip) of the rhesus monkey.
\newblock {\em Journal of neurophysiology}, 86(4):1916--1936, 2001.

\bibitem{kanwisher2006fusiform}
Nancy Kanwisher and Galit Yovel.
\newblock The fusiform face area: a cortical region specialized for the perception of faces.
\newblock {\em Philosophical Transactions of the Royal Society B: Biological Sciences}, 361(1476):2109--2128, 2006.

\bibitem{dehaene2011unique}
Stanislas Dehaene and Laurent Cohen.
\newblock The unique role of the visual word form area in reading.
\newblock {\em Trends in cognitive sciences}, 15(6):254--262, 2011.

\bibitem{mashour2020conscious}
George~A Mashour, Pieter Roelfsema, Jean-Pierre Changeux, and Stanislas Dehaene.
\newblock Conscious processing and the global neuronal workspace hypothesis.
\newblock {\em Neuron}, 105(5):776--798, 2020.

\bibitem{brown2020language}
Tom Brown, Benjamin Mann, Nick Ryder, Melanie Subbiah, Jared~D Kaplan, Prafulla Dhariwal, Arvind Neelakantan, Pranav Shyam, Girish Sastry, Amanda Askell, et~al.
\newblock Language models are few-shot learners.
\newblock {\em Advances in neural information processing systems}, 33:1877--1901, 2020.

\bibitem{grattafiori2024llama}
Aaron Grattafiori, Abhimanyu Dubey, Abhinav Jauhri, Abhinav Pandey, Abhishek Kadian, Ahmad Al-Dahle, Aiesha Letman, Akhil Mathur, Alan Schelten, Alex Vaughan, et~al.
\newblock The llama 3 herd of models.
\newblock {\em arXiv preprint arXiv:2407.21783}, 2024.

\bibitem{oquab2023dinov2}
Maxime Oquab, Timoth{\'e}e Darcet, Th{\'e}o Moutakanni, Huy Vo, Marc Szafraniec, Vasil Khalidov, Pierre Fernandez, Daniel Haziza, Francisco Massa, Alaaeldin El-Nouby, et~al.
\newblock Dinov2: Learning robust visual features without supervision.
\newblock {\em arXiv preprint arXiv:2304.07193}, 2023.

\bibitem{baevski2020wav2vec}
Alexei Baevski, Yuhao Zhou, Abdelrahman Mohamed, and Michael Auli.
\newblock wav2vec 2.0: A framework for self-supervised learning of speech representations.
\newblock {\em Advances in neural information processing systems}, 33:12449--12460, 2020.

\bibitem{chung2021w2v}
Yu-An Chung, Yu~Zhang, Wei Han, Chung-Cheng Chiu, James Qin, Ruoming Pang, and Yonghui Wu.
\newblock W2v-bert: Combining contrastive learning and masked language modeling for self-supervised speech pre-training.
\newblock In {\em 2021 IEEE Automatic Speech Recognition and Understanding Workshop (ASRU)}, pages 244--250. IEEE, 2021.

\bibitem{tong2022videomae}
Zhan Tong, Yibing Song, Jue Wang, and Limin Wang.
\newblock Videomae: Masked autoencoders are data-efficient learners for self-supervised video pre-training.
\newblock {\em Advances in neural information processing systems}, 35:10078--10093, 2022.

\bibitem{assran2025v}
Mido Assran, Adrien Bardes, David Fan, Quentin Garrido, Russell Howes, Matthew Muckley, Ammar Rizvi, Claire Roberts, Koustuv Sinha, Artem Zholus, et~al.
\newblock V-jepa 2: Self-supervised video models enable understanding, prediction and planning.
\newblock {\em arXiv preprint arXiv:2506.09985}, 2025.

\bibitem{yamins2014performance}
Daniel~LK Yamins, Ha~Hong, Charles~F Cadieu, Ethan~A Solomon, Darren Seibert, and James~J DiCarlo.
\newblock Performance-optimized hierarchical models predict neural responses in higher visual cortex.
\newblock {\em Proceedings of the national academy of sciences}, 111(23):8619--8624, 2014.

\bibitem{huth2016natural}
Alexander~G Huth, Wendy~A De~Heer, Thomas~L Griffiths, Fr{\'e}d{\'e}ric~E Theunissen, and Jack~L Gallant.
\newblock Natural speech reveals the semantic maps that tile human cerebral cortex.
\newblock {\em Nature}, 532(7600):453--458, 2016.

\bibitem{caucheteux2022brains}
Charlotte Caucheteux and Jean-R{\'e}mi King.
\newblock Brains and algorithms partially converge in natural language processing.
\newblock {\em Communications biology}, 5(1):134, 2022.

\bibitem{schrimpf2018brain}
Martin Schrimpf, Jonas Kubilius, Ha~Hong, Najib~J Majaj, Rishi Rajalingham, Elias~B Issa, Kohitij Kar, Pouya Bashivan, Jonathan Prescott-Roy, Franziska Geiger, et~al.
\newblock Brain-score: Which artificial neural network for object recognition is most brain-like?
\newblock {\em BioRxiv}, page 407007, 2018.

\bibitem{millet2022toward}
Juliette Millet, Charlotte Caucheteux, Yves Boubenec, Alexandre Gramfort, Ewan Dunbar, Christophe Pallier, Jean-Remi King, et~al.
\newblock Toward a realistic model of speech processing in the brain with self-supervised learning.
\newblock {\em Advances in Neural Information Processing Systems}, 35:33428--33443, 2022.

\bibitem{toneva2019interpreting}
Mariya Toneva and Leila Wehbe.
\newblock Interpreting and improving natural-language processing (in machines) with natural language-processing (in the brain).
\newblock {\em Advances in neural information processing systems}, 32, 2019.

\bibitem{linsley2025can}
Drew Linsley, Pinyuan Feng, and Thomas Serre.
\newblock Can deep neural networks learn biological vision?
\newblock {\em arXiv preprint arXiv:2504.16940}, 2025.

\bibitem{chehab2021deep}
Omar Chehab, Alexandre Defossez, Jean-Christophe Loiseau, Alexandre Gramfort, and Jean-Remi King.
\newblock Deep recurrent encoder: A scalable end-to-end network to model brain signals.
\newblock {\em arXiv preprint arXiv:2103.02339}, 2021.

\bibitem{hu2025neural}
Yu~Hu and Yalda Mohsenzadeh.
\newblock Neural processing of naturalistic audiovisual events in space and time.
\newblock {\em Communications Biology}, 8(1):110, 2025.

\bibitem{gao2023audiovisual}
Chuanji Gao, Jessica~J Green, Xuan Yang, Sewon Oh, Jongwan Kim, and Svetlana~V Shinkareva.
\newblock Audiovisual integration in the human brain: a coordinate-based meta-analysis.
\newblock {\em Cerebral Cortex}, 33(9):5574--5584, 2023.

\bibitem{beauchamp2005see}
Michael~S Beauchamp.
\newblock See me, hear me, touch me: multisensory integration in lateral occipital-temporal cortex.
\newblock {\em Current opinion in neurobiology}, 15(2):145--153, 2005.

\bibitem{driver2008multisensory}
Jon Driver and Toemme Noesselt.
\newblock Multisensory interplay reveals crossmodal influences on ‘sensory-specific’brain regions, neural responses, and judgments.
\newblock {\em Neuron}, 57(1):11--23, 2008.

\bibitem{stein2008multisensory}
Barry~E Stein and Terrence~R Stanford.
\newblock Multisensory integration: current issues from the perspective of the single neuron.
\newblock {\em Nature reviews neuroscience}, 9(4):255--266, 2008.

\bibitem{dahan2025sim}
Simon Dahan, Gabriel B{\'e}n{\'e}dict, Logan Zane~John Williams, Yourong Guo, Daniel Rueckert, Robert Leech, and Emma~Claire Robinson.
\newblock Sim: Surface-based fmri analysis for inter-subject multimodal decoding from movie-watching experiments.
\newblock In {\em The Thirteenth International Conference on Learning Representations}, 2025.

\bibitem{xia2024umbrae}
Weihao Xia, Raoul de~Charette, Cengiz Oztireli, and Jing-Hao Xue.
\newblock Umbrae: Unified multimodal brain decoding.
\newblock In {\em European Conference on Computer Vision}, pages 242--259. Springer, 2024.

\bibitem{gucclu2017modeling}
Umut G{\"u}{\c{c}}l{\"u} and Marcel~AJ Van~Gerven.
\newblock Modeling the dynamics of human brain activity with recurrent neural networks.
\newblock {\em Frontiers in computational neuroscience}, 11:7, 2017.

\bibitem{dong2023vision}
Dota~Tianai Dong and Mariya Toneva.
\newblock Vision-language integration in multimodal video transformers (partially) aligns with the brain.
\newblock {\em arXiv preprint arXiv:2311.07766}, 2023.

\bibitem{oota2022visio}
Subba~Reddy Oota, Jashn Arora, Vijay Rowtula, Manish Gupta, and Raju~S Bapi.
\newblock Visio-linguistic brain encoding.
\newblock {\em arXiv preprint arXiv:2204.08261}, 2022.

\bibitem{doerig2022semantic}
Adrien Doerig, Tim~C Kietzmann, Emily Allen, Yihan Wu, Thomas Naselaris, Kendrick Kay, and Ian Charest.
\newblock Semantic scene descriptions as an objective of human vision.
\newblock {\em arXiv preprint arXiv:2209.11737}, 10, 2022.

\bibitem{wang2022incorporating}
Aria~Y Wang, Kendrick Kay, Thomas Naselaris, Michael~J Tarr, and Leila Wehbe.
\newblock Incorporating natural language into vision models improves prediction and understanding of higher visual cortex.
\newblock {\em BioRxiv}, pages 2022--09, 2022.

\bibitem{tang2023brain}
Jerry Tang, Meng Du, Vy~Vo, Vasudev Lal, and Alexander Huth.
\newblock Brain encoding models based on multimodal transformers can transfer across language and vision.
\newblock {\em Advances in neural information processing systems}, 36:29654--29666, 2023.

\bibitem{jaegle2021perceiver}
Andrew Jaegle, Sebastian Borgeaud, Jean-Baptiste Alayrac, Carl Doersch, Catalin Ionescu, David Ding, Skanda Koppula, Daniel Zoran, Andrew Brock, Evan Shelhamer, Olivier H{\ifmmode\acute{e}\else\'{e}\fi}naff, Matthew~M. Botvinick, Andrew Zisserman, Oriol Vinyals, and Jo{\ifmmode\bar{a}\else\={a}\fi}o Carreira.
\newblock {Perceiver IO: A General Architecture for Structured Inputs {\&} Outputs}.
\newblock {\em arXiv}, July 2021.

\bibitem{srivastava2024omnivec}
Siddharth Srivastava and Gaurav Sharma.
\newblock Omnivec2 - a novel transformer based network for large scale multimodal and multitask learning.
\newblock In {\em Proceedings of the IEEE/CVF Conference on Computer Vision and Pattern Recognition (CVPR)}, pages 27412--27424, June 2024.

\bibitem{abdin2024phi4}
Marah Abdin, Jyoti Aneja, Harkirat Behl, S{\ifmmode\acute{e}\else\'{e}\fi}bastien Bubeck, Ronen Eldan, Suriya Gunasekar, Michael Harrison, Russell~J. Hewett, Mojan Javaheripi, Piero Kauffmann, James~R. Lee, Yin~Tat Lee, Yuanzhi Li, Weishung Liu, Caio C.~T. Mendes, Anh Nguyen, Eric Price, Gustavo de~Rosa, Olli Saarikivi, Adil Salim, Shital Shah, Xin Wang, Rachel Ward, Yue Wu, Dingli Yu, Cyril Zhang, and Yi~Zhang.
\newblock {Phi-4 Technical Report}.
\newblock {\em arXiv}, December 2024.

\bibitem{st2023cneuromod}
Marie St-Laurent, Basile Pinsard, Oliver Contier, Katja Seeliger, Valentina Borghesani, Julie Boyle, Pierre Bellec, and Martin Hebart.
\newblock cneuromod-things: a large-scale fmri dataset for task-and data-driven assessment of object representation and visual memory recognition in the human brain.
\newblock {\em Journal of Vision}, 23(9):5424--5424, 2023.

\bibitem{gifford2024algonauts}
Alessandro~T Gifford, Domenic Bersch, Marie St-Laurent, Basile Pinsard, Julie Boyle, Lune Bellec, Aude Oliva, Gemma Roig, and Radoslaw~M Cichy.
\newblock The algonauts project 2025 challenge: How the human brain makes sense of multimodal movies.
\newblock {\em arXiv preprint arXiv:2501.00504}, 2024.

\bibitem{esteban2019fmriprep}
Oscar Esteban, Christopher~J Markiewicz, Ross~W Blair, Craig~A Moodie, A~Ilkay Isik, Asier Erramuzpe, James~D Kent, Mathias Goncalves, Elizabeth DuPre, Madeleine Snyder, et~al.
\newblock fmriprep: a robust preprocessing pipeline for functional mri.
\newblock {\em Nature methods}, 16(1):111--116, 2019.

\bibitem{brett2002mni}
Matthew Brett.
\newblock The mni brain and the talairach atlas.
\newblock {\em www. mrc-Mrc-cbu. cam. ac. uk/Imaging/mnispace. Html}, 2002.

\bibitem{schaefer2018local}
Alexander Schaefer, Ru~Kong, Evan~M Gordon, Timothy~O Laumann, Xi-Nian Zuo, Avram~J Holmes, Simon~B Eickhoff, and BT~Thomas Yeo.
\newblock Local-global parcellation of the human cerebral cortex from intrinsic functional connectivity mri.
\newblock {\em Cerebral cortex}, 28(9):3095--3114, 2018.

\bibitem{wandell2011imaging}
Brian~A Wandell and Jonathan Winawer.
\newblock Imaging retinotopic maps in the human brain.
\newblock {\em Vision research}, 51(7):718--737, 2011.

\bibitem{defossez2023decoding}
Alexandre D{\'e}fossez, Charlotte Caucheteux, J{\'e}r{\'e}my Rapin, Ori Kabeli, and Jean-R{\'e}mi King.
\newblock Decoding speech perception from non-invasive brain recordings.
\newblock {\em Nature Machine Intelligence}, 5(10):1097--1107, 2023.

\bibitem{loshchilov2017decoupled}
Ilya Loshchilov and Frank Hutter.
\newblock Decoupled weight decay regularization.
\newblock {\em arXiv preprint arXiv:1711.05101}, 2017.

\bibitem{izmailov2018averaging}
Pavel Izmailov, Dmitrii Podoprikhin, Timur Garipov, Dmitry Vetrov, and Andrew~Gordon Wilson.
\newblock Averaging weights leads to wider optima and better generalization.
\newblock {\em arXiv preprint arXiv:1803.05407}, 2018.

\bibitem{jain2022hugging}
Shashank~Mohan Jain.
\newblock Hugging face.
\newblock In {\em Introduction to transformers for NLP: With the hugging face library and models to solve problems}, pages 51--67. Springer, 2022.

\bibitem{harris2020array}
Charles~R Harris, K~Jarrod Millman, St{\'e}fan~J Van Der~Walt, Ralf Gommers, Pauli Virtanen, David Cournapeau, Eric Wieser, Julian Taylor, Sebastian Berg, Nathaniel~J Smith, et~al.
\newblock Array programming with numpy.
\newblock {\em Nature}, 585(7825):357--362, 2020.

\bibitem{schoppe2016measuring}
Oliver Schoppe, Nicol~S. Harper, Ben D.~B. Willmore, Andrew~J. King, and Jan W.~H. Schnupp.
\newblock Measuring the performance of neural models.
\newblock {\em Frontiers in Computational Neuroscience}, Volume 10 - 2016, 2016.

\bibitem{antonello2023scaling}
Richard Antonello, Aditya Vaidya, and Alexander Huth.
\newblock Scaling laws for language encoding models in fmri.
\newblock {\em Advances in Neural Information Processing Systems}, 36:21895--21907, 2023.

\bibitem{jain2024computational}
Shailee Jain, Vy~A Vo, Leila Wehbe, and Alexander~G Huth.
\newblock Computational language modeling and the promise of in silico experimentation.
\newblock {\em Neurobiology of Language}, 5(1):80--106, 2024.

\end{thebibliography}

% \clearpage
\appendix
\section*{Appendices}
\section{Hyperparameters}
\label{app:hyperparameters}

\begin{table}[h]
\centering
\begin{tabular}{c|c|c}
\toprule
\textbf{Hyperparameter} & \textbf{Base value} & \textbf{Other values used for ensembling} \\
\midrule
Number of epochs & 15 \\
Number of TRs per window & 100 \\
Window jitter & 10s \\
Learning Rate & 10$^{-4}$ \\
Batch Size & 16 \\
Optimizer & AdamW \\
Scheduler type & OneCycleLR (cosine) \\
Scheduler warmup phase & 10\% \\
Stochastic weight average epochs & 8 \\
Dropout & 0 \\
Weight Decay & 0 \\
Hidden Size & 3072 \\
Text model & Llama-3.2-3B \\%& Qwen-2.5-1.5B \\
Audio model & Wav2Vec-Bert-2.0 \\%& Granite-Speech-3.3-2b\\
Video model & V-JEPA-2-Gigantic-256 \\%& VideoMAE-Huge \\
\midrule
% Features to use & [text, audio, video] & [text, audio], [audio, video], [video, text] \\
Loss & MSE & Pearson, SmoothL1, HuberLoss \\
Modality Dropout & 0.2 & 0.0, 0.4 \\
Layer groups & [0.5, 0.75, 1] & [0,0.5,1], [0.5, 1], [0, 0.2, 0.4, 0.6, 0.8, 1.]\\
Layer extraction & group by intervals & extract single layers \\
Layer aggregation & concatenate & average \\
Modality aggregation & concatenate & average \\
Use subject embedding & \cmark & \xmark \\
\bottomrule
\end{tabular}
\caption{Hyperparameters used in our model}
\label{tab:hyperparameters}
\end{table}

\section{Licenses}
\label{app:licenses}

HuggingFace models:
\begin{itemize}
\item Video-JEPA 2: Apache (\url{https://github.com/facebookresearch/vjepa2/blob/main/LICENSE})
    \item Wav2Vec-Bert-2.0: MIT (\url{https://huggingface.co/datasets/choosealicense/licenses/blob/main/markdown/mit.md})
    \item LLama-3.2-3B: llama3.2 (\url{https://huggingface.co/meta-llama/Llama-3.2-1B/blob/main/LICENSE.txt})
\end{itemize}

Packages:
\begin{itemize}
    \item \texttt{x-transformers}: MIT (\url{https://github.com/lucidrains/x-transformers/blob/main/LICENSE})
    \item \texttt{nilearn}:  BSD (\url{https://github.com/nilearn/nilearn/blob/main/LICENSE})
    \item \texttt{pytorch}: \url{https://github.com/pytorch/pytorch/blob/main/LICENSE}
\end{itemize}

Datasets:
\begin{itemize}
    \item Courtois NeuroMod: CC0 (\url{https://creativecommons.org/publicdomain/zero/1.0/legalcode})
\end{itemize}

% \begin{figure}
%     \centering
%     \includegraphics[width=0.8\linewidth]{figs/topomap_linear_vs_deep.pdf}
%     \caption{\textbf{Performance of deep encoding model compared to linear encoding model.} }
%     \label{fig:linear-vs-deep}
% \end{figure}

% \begin{figure}
%     \centering
%     \includegraphics[width=0.5\linewidth]{figs/lebel2_atlas_dim_acc.pdf}
%     \caption{Dependence on atlas dimensionality}
%     \label{fig:atlas-dim}
% \end{figure}

\newpage

\end{document}